\documentclass[a4paper,fleqn]{cas-sc}

\usepackage[authoryear,longnamesfirst]{natbib}

\usepackage{graphicx} 
\usepackage{amsmath, amssymb} 
\usepackage{hyperref} 

\begin{document}

\shorttitle{Multimodal Deep Learning Framework for Edema Classification}
\title[mode=title]{A Multimodal Deep Learning Framework for Edema Classification Using HCT and Clinical Data}

\author[1]{Aram Ansary Ogholbake}
\affiliation[1]{organization={Department of Internal Medicine and Department of Computer Science, University of Kentucky},
                city={Lexington},
                state={KY},
                country={USA}}
\credit{Conceptualization, Investigation, Methodology, Software, Visualization, Writing – original draft, Writing – review and editing}

\author[2]{Hannah Choi}
\credit{Data curation, Writing – review and editing}
\author[3]{Spencer Brandenburg}
\credit{Data curation, Writing – review and editing}
\author[3]{Alyssa Antuna}
\credit{Data curation, Writing – review and editing}
\author[2]{Zahraa Al-Sharshahi}
\credit{Data curation, Writing – review and editing}
\author[3]{Makayla Cox}
\credit{Data curation, Writing – review and editing}
\author[3]{Haseeb Ahmed}
\credit{Data curation, Writing – review and editing}
\author[2]{Jacqueline Frank}
\credit{Data curation, Writing – review and editing}
\author[2]{Nathan Millson}
\credit{Data curation, Writing – review and editing}
\author[2]{Luke Bauerle}
\credit{Data curation, Writing – review and editing}
\author[4]{Jessica Lee}
\credit{conceptualization, funding acquisition, methodology, Writing – review and editing}
\author[2]{David Dornbos III}
\credit{conceptualization, funding acquisition, methodology, Writing – review and editing}
\author[1]{Qiang Cheng}[orcid=0000-0002-3596-2838]
\cormark[1]
\ead{qiang.cheng@uky.edu}
\credit{Conceptualization, Funding acquisition, Investigation, Methodology, Project administration, Supervision, Writing – review and editing}

\shortauthors{Ansary Ogholbake et~al.}

\affiliation[2]{organization={Department of Neurological Surgery, University of Kentucky},
                city={Lexington},
                state={KY},
                country={USA}}

\affiliation[3]{organization={University of Kentucky College of Medicine},
                city={Lexington},
                state={KY},
                country={USA}}

\affiliation[4]{organization={Department of Neurology, University of Kentucky},
                city={Lexington},
                state={KY},
                country={USA}}

\cortext[cor1]{Corresponding author}

\begin{abstract}
We propose AttentionMixer, a unified deep learning framework for multimodal detection of brain edema that combines structural head CT (HCT) with routine clinical metadata. While HCT provides rich spatial information, clinical variables such as age, laboratory values, and scan timing capture complementary context that might be ignored or naively concatenated. AttentionMixer is designed to fuse these heterogeneous sources in a principled and efficient manner. HCT volumes are first encoded using a self-supervised Vision Transformer Autoencoder (ViT-AE++), without requiring large labeled datasets. Clinical metadata are mapped into the same feature space and used as keys and values in a cross-attention module, where HCT-derived feature vector serves as queries. This cross-attention fusion allows the network to dynamically modulate imaging features based on patient-specific context and provides an interpretable mechanism for multimodal integration. A lightweight MLP-Mixer then refines the fused representation before final classification, enabling global dependency modeling with substantially reduced parameter overhead. Missing or incomplete metadata are handled via a learnable embedding, promoting robustness to real-world clinical data quality. We evaluate AttentionMixer on a curated brain HCT cohort with expert edema annotations using five-fold cross-validation. Compared with strong HCT-only, metadata-only, and prior multimodal baselines, AttentionMixer achieves superior performance (accuracy 87.32\%, precision 92.10\%, F1-score 85.37\%, AUC 94.14\%). Ablation studies confirm the benefit of both cross-attention and MLP-Mixer refinement, and permutation-based metadata importance analysis highlights clinically meaningful variables driving predictions. These results demonstrate that structured, interpretable multimodal fusion can substantially improve edema detection in clinical practice.
\end{abstract}

\begin{graphicalabstract}
\includegraphics{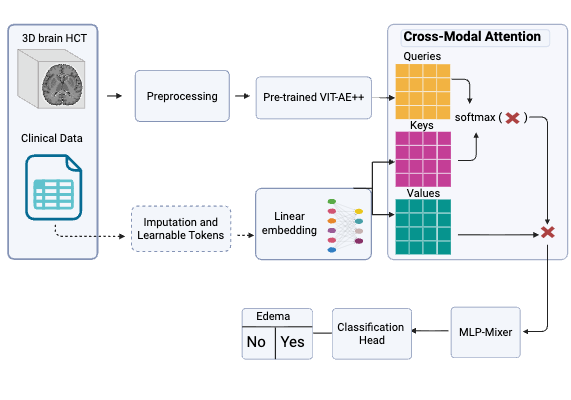}
\end{graphicalabstract}

\begin{highlights}

\item Propose AttentionMixer which detects edema using multimodal head CT and clinical metadata.
\item Handles missing clinical data with a learnable metadata embedding.
\item Combines cross-attention and MLP-Mixer in its multimodal framework.
\item Outperforms strong unimodal and multimodal baselines.
\end{highlights}

\maketitle

\section{Introduction}
Brain edema is a common and potentially life-threatening complication of acute neurologic conditions such as ischemic stroke, traumatic brain injury, and brain tumors. It contributes to elevated intracranial pressure, brain herniation, and poor functional outcomes, making timely detection and monitoring crucial for triage, treatment planning, and prognostication \cite{stokum2016molecular,koenig2018cerebral}. In current clinical practice, edema assessment largely relies on radiologists or neurosurgeons visually inspecting head CT (HCT) scans, a process that is time-consuming, subject to inter-observer variability, and prone to missing subtle or diffuse patterns, especially in high-volume workflows.

Deep learning has substantially advanced automated analysis of brain HCT, particularly for tumor and lesion segmentation. Convolutional and transformer-based architectures have been developed and benchmarked on public datasets such as BraTS to delineate tumors and associated peritumoral edema \cite{menze2015brats}. Models such as DeepMedic and related 3D CNN frameworks have demonstrated strong performance for brain lesion segmentation across a range of pathologies \cite{kamnitsas2017efficient}. More recently, deep learning has been applied to automated measurement of hemorrhage and perihematomal edema and other edema-related imaging endpoints \cite{dhar2020deep}. However, many of these approaches focus on voxel-wise segmentation, require extensive pixel-level annotations, and are often tailored to radiotherapy planning or research settings rather than robust, patient-level edema detection suitable for deployment as a screening or decision-support tool.

In routine clinical decision-making, HCT interpretation is rarely performed in isolation. Clinical factors such as age, comorbidities, laboratory values, and scan timing strongly influence whether ambiguous signal abnormalities are interpreted as vasogenic or cytotoxic edema, rather than other pathologies. Multimodal machine learning studies in stroke and related neurologic disease show that integrating neuroimaging with structured clinical variables can improve functional outcome prediction and risk stratification compared with imaging-only models \cite{jo2023combining}. Despite this, multimodal architectures that explicitly fuse 3D brain HCT with tabular clinical metadata for edema detection remain scarce.

More broadly, multimodal biomedical machine learning faces several challenges. Imaging and non-imaging features live on different scales, clinical variables are frequently missing or noisy, and the relative importance of each modality can vary across patients and pathologies. Many existing multimodal methods resort to simple early fusion \cite{snoek2005early, zhao2024deep} (simple feature concatenation) or late fusion \cite{morvant2014majority} (combining modality-specific subnetworks only at the classifier), which limits their ability to model fine-grained interactions between modalities and complicates interpretability \cite{cui2023deep,li2024review,stahlschmidt2022multimodal}. Recent work on attention-based multimodal fusion and robust prediction with missing modalities suggests that more principled fusion mechanisms, particularly attention-based ones, can better handle heterogeneity and missingness in clinical settings \cite{stahlschmidt2022multimodal,liu2023armour}. However, most such methods are designed for 2D images, non-brain applications, or text–image pairs, and do not directly address fusion of 3D brain HCT with structured clinical metadata for edema detection.

In parallel, transformer-based vision models and self-supervised learning have emerged as powerful tools for representation learning in medical imaging. Vision Transformers (ViT) and masked autoencoders (MAE) learn high-level image representations by treating images as sequences of patches and reconstructing masked inputs \cite{dosovitskiy2021vit,he2022mae}. Self-supervised learning has been shown to improve data efficiency and downstream performance for various medical image classification and segmentation tasks, particularly when labeled data are limited \cite{huang2023selfsupervised,chen2023mim3d}. These advances suggest that ViT-style autoencoders are well-suited for extracting rich features from brain HCT scans in cohorts where voxel-wise edema annotations are scarce. Yet, it remains unclear how to best combine such powerful image encoders with structured patient metadata in a way that is parameter-efficient, robust to missing values, and clinically interpretable.

In this work, we address these gaps by proposing \emph{AttentionMixer}, a unified multimodal deep learning framework for edema detection from 3D brain HCT and routine clinical metadata. AttentionMixer first encodes HCT scans using a self-supervised ViT-based autoencoder to obtain a high-quality feature extraction, while clinical variables are mapped into the same latent space via a lightweight embedding network. A cross-attention module \cite{abavisani2020multimodal} then treats HCT features as queries and metadata embeddings as keys and values, enabling the model to adaptively modulate imaging features based on patient-specific clinical context and providing a transparent mechanism for multimodal interaction.  To refine the fused representation with minimal computational overhead, we leverage an MLP-Mixer module, which does channel-mixing MLPs to capture global dependencies in a parameter-efficient manner \cite{tolstikhin2021mlp}. Finally, to cope with incomplete real-world metadata, we introduce a learnable missing-value embedding that substitutes entirely missing clinical vectors, enabling training and inference without discarding patients with incomplete records and aligning with recent attention-based strategies for robust multimodal prediction under missingness \cite{liu2023armour}.

We evaluate AttentionMixer on a curated cohort of patients with expert-annotated edema labels, comparing against strong HCT-only, metadata-only, and multimodal baselines, and conducting ablation studies on fusion strategies. Our model consistently outperforms all alternatives, achieving higher accuracy, F1-score, and area under the ROC curve, and our permutation-based feature importance analysis highlights clinically meaningful metadata that drive the cross-attention modulation of HCT features.

Our key contributions are summarized as follows:
\begin{itemize}
    \item A unified multimodal framework for edema detection.
    We propose AttentionMixer, an end-to-end architecture that jointly leverages 3D brain HCT and structured clinical metadata for patient-level edema classification, moving beyond purely imaging-based or voxel-wise segmentation approaches.
    
    \item Principled and interpretable fusion via cross-attention.
    We design a cross-attention module that uses HCT features as queries and metadata as keys/values, enabling patient-specific modulation of imaging representations and offering a transparent mechanism to analyze how clinical factors influence image-based predictions.
    
    \item Efficient post-fusion refinement with an MLP-Mixer.
    Instead of stacking additional transformer layers, we employ an MLP-Mixer for post-fusion refinement, retaining the ability to model global dependencies while being substantially more parameter-efficient and thus more amenable to clinical deployment \cite{tolstikhin2021mlp}.
    
    \item Self-supervised ViT-based encoder for HCT with limited labels.
    We adapt a self-supervised ViT-style encoder to extract high-level HCT representations, leveraging masked image modeling pretraining to reduce overfitting in small medical cohorts and to improve downstream edema classification performance \cite{dosovitskiy2021vit,he2022mae,huang2023selfsupervised,chen2023mim3d}.
    
    \item Robust metadata handling with a learnable missing-value embedding.
    We explicitly model missing clinical records through a trainable embedding token that replaces entirely missing metadata vectors, enabling robust training and inference under realistic patterns of clinical missingness and aligning with recent attention-based multimodal methods designed to handle missing modalities \cite{stahlschmidt2022multimodal,liu2023armour}.
\end{itemize}

\section{Related Work}
In the following, we provide a brief review of the related work in the literature. 

\textbf{Automated edema and lesion analysis on brain imaging.}
Deep learning has been widely applied to automated analysis of brain HCT, especially for brain tumor and lesion segmentation. The BraTS benchmark popularized multi-sequence MRI segmentation of tumors and peritumoral edema, spurring the development of 3D CNN and transformer-based architectures that can closely approximate expert delineations \cite{menze2015brats}. Methods such as DeepMedic and related multi-scale 3D CNN frameworks combine patch-based processing with conditional random fields or dense CRF post-processing to achieve robust lesion segmentation across diverse pathologies \cite{kamnitsas2017efficient}. Beyond tumors, deep networks and machine-learning models have been applied to edema-related tasks, including automated volumetric quantification of perihematomal edema on HCT \cite{dhar2020deep} and radiomics-based prediction of perihematomal edema from CT \cite{chen2023machine}, as well as  machine learning methods using near-infrared spectroscopy \cite{shah2023advancing}. However, these approaches primarily focus on \textit{voxel-wise segmentation} or handcrafted radiomics pipelines, require labor-intensive pixel-level annotation or feature engineering, and rarely incorporate structured clinical metadata. Moreover, they are typically optimized for imaging biomarkers (e.g., lesion) rather than patient-level HCT-based edema detection suitable for screening.

\textbf{Fusion of imaging and clinical metadata.}
A growing body of work explores multimodal fusion of medical images with non-image data such as demographics, comorbidities, laboratory values, and other electronic health record (EHR) features. For example, combining clinical and imaging data has been shown to improve functional outcome prediction after acute ischemic stroke relative to models using either modality alone \cite{jo2023combining}. Several reviews provide taxonomies of multimodal fusion strategies in biomedical applications, distinguishing early fusion (feature concatenation), late fusion (ensemble of modality-specific predictors), and intermediate or joint fusion strategies that learn shared latent representations \cite{cui2023deep,stahlschmidt2022multimodal,li2024review}. These surveys highlight the limitations of naïve early and late fusion, particularly in terms of capturing fine-grained cross-modal interactions and dealing with heterogeneous scaling and missingness in clinical variables.

More recently, methods have begun to explicitly condition image processing on tabular data. HyperFusion \cite{duenias2025hyperfusion} introduced a hypernetwork-based framework in which tabular EHR features generate parameters for a primary CNN operating on brain MRI, enabling the imaging pipeline to adapt to patient-specific clinical context. HyperFusion demonstrated improved performance over unimodal and simpler multimodal baselines on brain age prediction and Alzheimer’s disease classification. Our work is conceptually related in that we also integrate 3D brain HCT with tabular clinical data, but differs in several key aspects: (i) we target binary edema detection rather than regression or multi-class diagnosis, (ii) we rely on cross-attention at the feature level instead of hypernetwork-generated weights, and (iii) we explicitly combine self-supervised ViT-based HCT encoders with cross-attention and a lightweight MLP-Mixer for post-fusion refinement.

\textbf{Multimodal attention and robustness to missing modalities.}
Attention mechanisms have become popular for modeling interactions between modalities in biomedical applications. Taxonomies of multimodal biomedical fusion explicitly identify attention-based intermediate fusion as a powerful strategy for capturing cross-modal dependencies and enabling more interpretable models \cite{amador2025cross,stahlschmidt2022multimodal}. Liu et al.\ propose an attention-based multimodal fusion framework with contrastive learning that is explicitly designed to be robust when some modalities are missing at training or inference time, showing improved clinical prediction performance in realistic EHR settings \cite{liu2023armour}. While these approaches underscore the benefits of attention for multimodal learning and robustness to missing modalities, most existing work focuses on 2D images, text, or EHR-only data, rather than 3D brain HCT fused with structured clinical metadata. Furthermore, missingness is often handled at the modality level (i.e., entire modalities absent), whereas real-world clinical practice frequently involves partially missing or noisy metadata fields.

\textbf{Self-supervised vision transformers for medical imaging.}
Transformers and self-supervised learning have significantly advanced representation learning in medical imaging. Vision Transformers (ViT) treat images as sequences of patches, enabling global context modeling and competitive performance on large-scale benchmarks \cite{dosovitskiy2021vit}. Masked autoencoders (MAE) further showed that reconstructing masked patches is an effective and scalable self-supervised objective for images \cite{he2022mae}. Systematic reviews of self-supervised learning in medical imaging highlight that such pretraining substantially improves downstream classification and segmentation, particularly under limited annotation budgets \cite{huang2023selfsupervised}. Masked image modeling has been adapted to 3D medical images, including CT and MRI, where pretraining on unlabeled volumes yields better performance and data efficiency for various downstream tasks \cite{chen2023mim3d}. These results motivate our choice to build on a ViT-style autoencoder to extract high-level brain HCT representations for edema detection, alleviating the need for large voxel-wise labeled datasets.

\textbf{MLP-Mixer and efficient architectures.}
The MLP-Mixer architecture replaces convolutions and self-attention with alternating token-mixing and channel-mixing MLPs applied to patch embeddings \cite{tolstikhin2021mlp}. Despite its simplicity, MLP-Mixer achieves competitive performance to transformers and CNNs on standard image benchmarks while offering a favorable trade-off between accuracy and parameter efficiency. In medical imaging, similar ideas—using pure MLP blocks or mixer-based modules—have been explored to reduce computational cost and simplify deployment, but have rarely been combined with self-supervised ViT encoders and multimodal fusion \cite{yang2025d2,zhang2023improving}. In our work, we use an MLP-Mixer not as a standalone vision backbone but as a lightweight post-fusion refinement module, allowing global interactions within the fused HCT–metadata representation while keeping the overall model compact and more suitable for clinical deployment.

\textbf{Summary.}
In summary, prior work has (i) demonstrated strong deep-learning performance for tumor and edema-related lesion segmentation on brain imaging \cite{menze2015brats,kamnitsas2017efficient,dhar2020deep,chen2023machine,shah2023advancing}, (ii) shown that integrating clinical and imaging modalities can improve prediction tasks \cite{jo2023combining,cui2023deep,stahlschmidt2022multimodal,li2024review,amador2025cross}, (iii) introduced attention-based fusion and hypernetwork-based approaches such as HyperFusion for imaging–tabular integration \cite{liu2023armour,duenias2025hyperfusion}, and (iv) established self-supervised ViT and mixer-style architectures as powerful yet efficient visual backbones \cite{dosovitskiy2021vit,he2022mae,huang2023selfsupervised,chen2023mim3d,tolstikhin2021mlp,yang2025d2,zhang2023improving}. AttentionMixer is distinguished by unifying these directions in a single architecture tailored for brain edema detection: it uses a self-supervised ViT-based encoder for 3D HCT, cross-attention to condition imaging features on clinical metadata, an MLP-Mixer for parameter-efficient post-fusion refinement, and an explicit learnable embedding to handle entirely missing metadata at the patient level.

\section{Method}
\begin{figure*}[!ht]
    \centering
    \includegraphics[width=\textwidth]{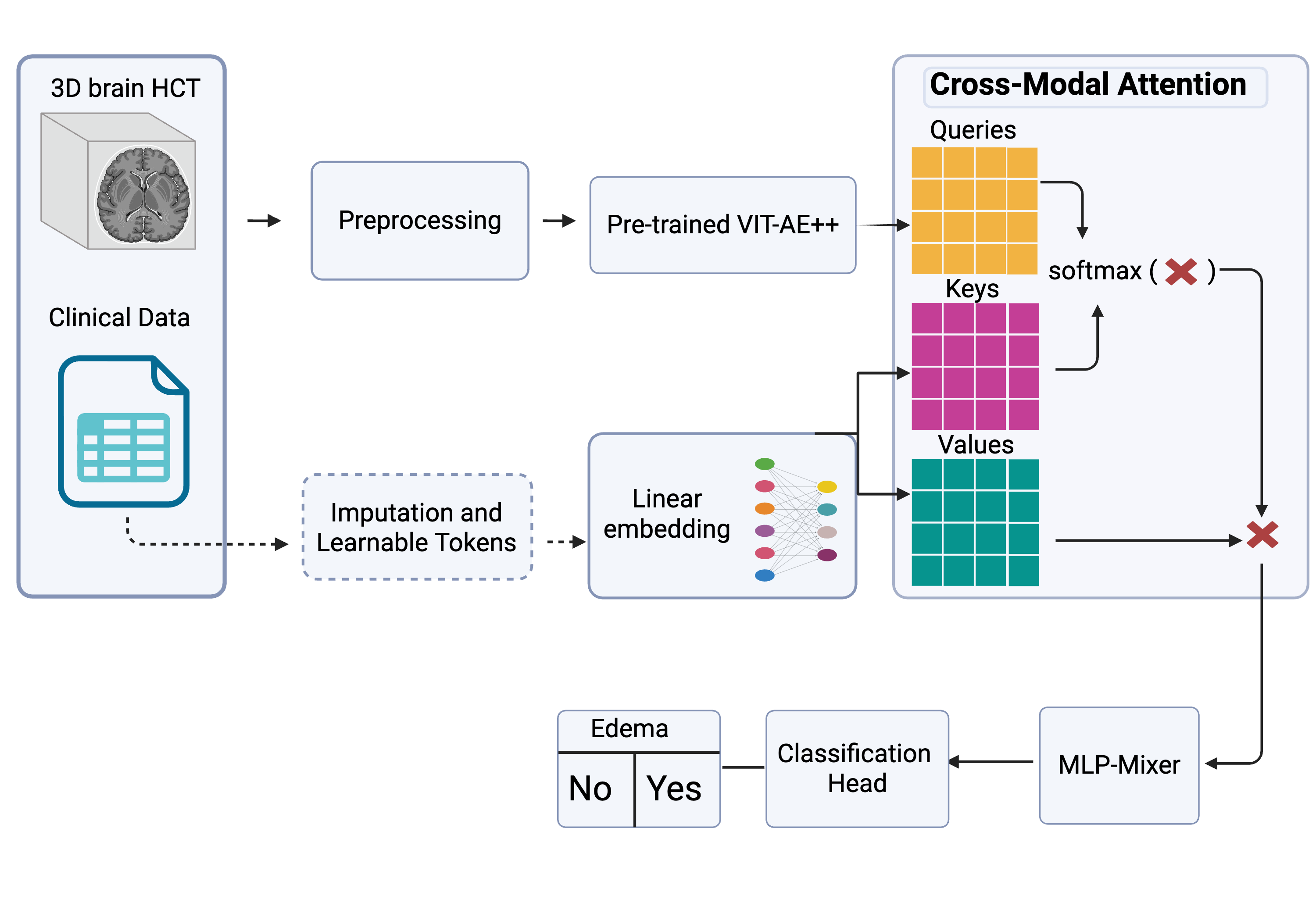}
    \caption{Overall diagram of AttentionMixer. HCT volumes are encoded by a self-supervised ViT-based encoder, while tabular metadata are embedded into the same latent space. A cross-attention module uses HCT features as queries and metadata as keys/values to produce a fused representation, which is further refined by an MLP-Mixer before classification.}
    \label{fig:overall_architecture}
\end{figure*}

In this section, we formally describe \emph{AttentionMixer}, a multimodal learning framework that fuses 3D brain HCT and clinical metadata for edema classification at the patient level. The overall architecture is shown in Figure~\ref{fig:overall_architecture}.

At a high level, AttentionMixer comprises four stages:
\begin{enumerate}
    \item Pre-processing of HCT volumes and clinical metadata;
    \item Independent feature extraction from HCT and metadata, using a ViT-based encoder for imaging and a linear embedding network for tabular data;
    \item Cross-modal feature alignment via a cross-attention module that conditions HCT features on patient-specific metadata; and
    \item Post-fusion refinement and classification using an MLP-Mixer followed by a linear classifier.
\end{enumerate}

Cross-attention and the MLP-Mixer play complementary roles in this pipeline. The cross-attention module implements a principled fusion mechanism by treating HCT-derived tokens as queries and metadata embeddings as keys and values. This allows the network to dynamically modulate imaging features based on clinical context (e.g., age, lab values, scan timing), rather than simply concatenating modalities or processing them independently. However, cross-attention by itself primarily focuses on aligning information across modalities and does not fully capture higher-order interactions within the fused representation. To address this, we introduce an MLP-Mixer as a lightweight yet expressive refinement layer. By using channel-mixing MLPs on the fused features, the MLP-Mixer models global dependencies in a parameter-efficient way, supporting both task-specific multimodal fusion and deeper interaction modeling while remaining suitable for clinical deployment.

\subsection{Problem Formulation}
Let $\mathcal{D} = \{(X_i, M_i, y_i)\}_{i=1}^{N}$ denote a dataset of $N$ patients, where each sample consists of an HCT volume $X_i \in \mathbb{R}^{D \times H \times W}$, a tabular metadata vector $M_i \in \mathbb{R}^{d_m}$, and a corresponding binary edema label $y_i \in \{0,1\}$. Here, $D$ denotes the number of slices (or depth), $H \times W$ denotes the in-plane spatial resolution, and $d_m$ denotes the number of variables in the metadata.

Our goal is to learn a function
\[
    f: \bigl(\mathbb{R}^{D \times H \times W}, \mathbb{R}^{d_m}\bigr) \rightarrow \{0,1\}
\]
that predicts the presence or absence of edema by leveraging both modalities. We parameterize $f$ as a composition of modules:
\begin{equation}
    f(X, M) = g_{\text{cls}} \circ h_{\text{mix}} \circ \phi_{\text{cross}} \circ \bigl(f_{\text{HCT}}(X), f_{\text{meta}}(M)\bigr),
\end{equation}
where:
\begin{itemize}
    \item $f_{\text{HCT}}: \mathbb{R}^{D \times H \times W} \rightarrow \mathbb{R}^{d_f}$ is an encoder that extracts deep features from HCT (initialized from a self-supervised ViT-based autoencoder).
    \item $f_{\text{meta}}: \mathbb{R}^{d_m} \rightarrow \mathbb{R}^{d_f}$ maps tabular metadata into a shared latent feature space of dimension $d_f$.
    \item $\phi_{\text{cross}}: (\mathbb{R}^{d_f}, \mathbb{R}^{d_f}) \rightarrow \mathbb{R}^{d_f}$ fuses HCT and metadata features via cross-attention.
    \item $h_{\text{mix}}: \mathbb{R}^{d_f} \rightarrow \mathbb{R}^{d_f}$ applies an MLP-Mixer to refine the fused representation.
    \item $g_{\text{cls}}: \mathbb{R}^{d_f} \rightarrow \{0, 1\}$ is the final classifier producing the edema prediction.
\end{itemize}

In the following subsections, we detail each component, including data pre-processing, HCT feature extraction with a ViT-based autoencoder, metadata embedding and handling of missing values, cross-modal attention fusion, and MLP-Mixer-based refinement and classification.

\subsection{Preprocessing of HCT and Metadata}

We first standardize both HCT volumes and clinical metadata to obtain stable inputs for AttentionMixer.  
For each HCT scan $X_i$, we apply (i) isotropic resampling to a fixed resolution of $128^3$ and (ii) intensity normalization . Specifically, all volumes are resampled to an isotropic voxel spacing, yielding $X_i \in \mathbb{R}^{D \times H \times W}$ with $D = H = W = 128$. Then voxel intensities are clipped to the 1st and 99th percentiles, and linearly rescaled to the $[0, 1]$ range to reduce the influence of extreme outliers and scanner-dependent variations. 

Clinical metadata $M_i \in \mathbb{R}^{d_m}$ are preprocessed using standard tabular pipelines.  Categorical variables are one-hot encoded. Missing or corrupted entries at the feature level are imputed using a $k$-nearest neighbors ($k$-NN) imputer \cite{troyanskaya2001missing} fitted on the training set, providing a complete metadata vector for most patients.

In practice, some patients may lack an entire metadata record (e.g., no tabular information available in the EHR for that encounter). For these cases, rather than discarding samples, we replace the metadata vector $M_i$ with a learnable embedding $\mathbf{e}_m \in \mathbb{R}^{d_m}$ at the input of the metadata encoder. After passing through the metadata embedding network (Section~\ref{sec:metadata_embedding}), $\mathbf{e}_m$ is mapped into the latent feature space $\mathbb{R}^{d_f}$, allowing the model to learn how to handle patients with completely missing metadata in a data-driven manner.


\begin{table*}[ht]
\centering
\renewcommand{\arraystretch}{1.6}
\caption{Performance comparison of edema classification models.}
\label{tab:main_comparison}
\resizebox{\textwidth}{!}{%
\begin{tabular}{|l|c|c|c|c|c|}
\hline
\textbf{Model} & \textbf{Modality} & \textbf{Accuracy (\%)} & \textbf{Precision (\%)} & \textbf{F1-Score (\%)} & \textbf{AUC (\%)} \\
\hline
3D CNN \cite{tran2015learning}& HCT only & 78.68 & 86.69 & 76.49 & 84.00 \\
DenseNet \cite{huang2017densely} & HCT only & 77.07 & 81.05 & 72.21 & 79.95 \\
ViT \cite{vaswani2017attention} & HCT only & 70.24 & 61.34 & 55.92 & 73.35 \\
MLP-Mixer \cite{tolstikhin2021mlp} & HCT only & 74.63 & 74.79 & 71.67 & 76.57 \\
MLP & Metadata only & 84.88 & 89.90 & 82.77 & 92.31 \\
\hline
HyperFusion \cite{duenias2025hyperfusion} & HCT + metadata & 66.66 & 73.33 & 72.72 & 86.66 \\
\textbf{AttentionMixer (Ours)} & HCT + metadata & \textbf{87.32} & \textbf{92.10} & \textbf{85.37} & \textbf{94.14} \\
\hline
\end{tabular}%
}
\end{table*}

\begin{table*}[ht]
\centering
\renewcommand{\arraystretch}{1.6}
\caption{Ablation study: impact of fusion strategies on edema classification. CA denotes cross-attention.}
\label{tab:ablation_fusion}
\resizebox{\textwidth}{!}{%
\begin{tabular}{|l|c|c|c|c|c|}
\hline
\textbf{Fusion Method} & \textbf{Attention Type}  & \textbf{Accuracy (\%)} & \textbf{Precision (\%)} & \textbf{F1-Score (\%)} & \textbf{AUC (\%)} \\
\hline
AttentionMixer (w/o CA) & None & 86.83 & 87.92 & 85.55 & 92.66 \\
AttentionMixer (w/o MLP-Mixer) & Cross-attention only & 85.85 & 92.66 & 84.04 & 93.35 \\
Early Fusion & None & 85.37 & 84.45 & 83.95 & 90.00 \\
Only MLP-Mixer & None & 70.00 & 87.32 & 67.31 & 65.00 \\
\hline
\end{tabular}%
}
\end{table*}

\subsection{HCT Feature Extraction with Vision Transformer Autoencoder}
\label{sec:HCT_encoder}
To extract high-level feature representations from HCT scans, we employ a pretrained Vision Transformer autoencoder (ViT-AE++) \cite{prabhakar2024vit}, initialized from a self-supervised masked image modeling objective. The encoder $f_{\text{HCT}}$ operates on 3D HCT volumes and decomposes each scan into a sequence of non-overlapping patches.

Formally, given a preprocessed HCT volume $X_i \in \mathbb{R}^{D \times H \times W}$, we partition it into non-overlapping 3D patches and processes them through stacked transformer blocks, producing a latent feature volume
\begin{equation}
    F_{\text{HCT}} = f_{\text{HCT}}(X_i) \in \mathbb{R}^{C \times D' \times H' \times W'},
\end{equation}
where $C$ denotes the number of latent feature channels and $D', H', W'$ are the
spatial dimensions of the encoded representation.

Since downstream components operate in a unified latent dimension $d_f$, we flatten
the encoder output and apply a linear projection to obtain a compact HCT feature representation:
\begin{equation}
    F_{\text{HCT}}' = W_{\text{HCT}} \, \mathrm{vec}(F_{\text{HCT}}) \in \mathbb{R}^{d_f},
\end{equation}
where $\mathrm{vec}(\cdot)$ denotes vectorization of the 3D feature volume
$F_{\text{HCT}} \in \mathbb{R}^{C \times D' \times H' \times W'}$, and
$W_{\text{HCT}} \in \mathbb{R}^{d_f \times (C D' H' W')}$ is a learnable projection
matrix. This produces a single latent vector $F_{\text{HCT}}'$ of dimension $d_f$. For use in the cross-attention module, we reshape the HCT embedding into a single-token sequence. We reuse the notation $F_{\text{HCT}}' \in \mathbb{R}^{1 \times d_f}$ below.

\subsection{Metadata Processing and Embedding}
\label{sec:metadata_embedding}

The tabular metadata $M_i \in \mathbb{R}^{d_m}$, after preprocessing and feature-level imputation, are mapped into the same latent space as the HCT features via a linear embedding network:
\begin{equation}
    F_{\text{meta}}' = f_{\text{meta}}(M_i) = W_{\text{meta}} M_i + b_{\text{meta}} \in \mathbb{R}^{d_f},
\end{equation}
where $W_{\text{meta}} \in \mathbb{R}^{d_f \times d_m}$ and $b_{\text{meta}} \in \mathbb{R}^{d_f}$ are learnable parameters. When an entire metadata vector is missing for a patient, we replace $M_i$ with the learnable embedding $\mathbf{e}_m$ at the input, so that $f_{\text{meta}}(\mathbf{e}_m)$ becomes the metadata representation. This design allows the network to learn a principled default representation for missing clinical records, instead of relying on ad hoc imputation rules.

For cross-attention, we interpret $F_{\text{meta}}'$ as a (possibly single-token) sequence in $\mathbb{R}^{L \times d_f}$, where $L$ is the number of metadata tokens (typically $L=1$). We reuse the notation $F_{\text{meta}}' \in \mathbb{R}^{L \times d_f}$ below.

\subsection{Cross-Modal Attention Fusion}

To model interactions between HCT-derived features and clinical metadata, we employ a cross-modal attention mechanism that conditions imaging tokens on patient-specific metadata. Let $F_{\text{HCT}}' \in \mathbb{R}^{1 \times d_f}$ denote the HCT embedding and $F_{\text{meta}}' \in \mathbb{R}^{L \times d_f}$ represent the corresponding metadata embeddings, with $L = 1$ in the standard single-token setting. The cross-attention module $\phi_{\text{cross}}$ produces a fused representation:
\begin{equation}
    F_{\text{fused}} = \phi_{\text{cross}}\bigl(F_{\text{HCT}}', F_{\text{meta}}'\bigr) \in \mathbb{R}^{1 \times d_f}.
\end{equation}

\paragraph{Multi-head cross-attention.}
We implement $\phi_{\text{cross}}$ using standard multi-head scaled dot-product attention, where HCT tokens serve as queries and metadata tokens act as keys and values. For a single head, we compute:
\begin{align}
    Q &= F_{\text{HCT}}' W_Q, \quad Q \in \mathbb{R}^{1 \times d_k}, \\
    K &= F_{\text{meta}}' W_K, \quad K \in \mathbb{R}^{1 \times d_k}, \\
    V &= F_{\text{meta}}' W_V, \quad V \in \mathbb{R}^{1 \times d_v},
\end{align}
with learnable projection matrices $W_Q, W_K \in \mathbb{R}^{d_f \times d_k}$ and $W_V \in \mathbb{R}^{d_f \times d_v}$.

The attention weights between HCT token $i$ and metadata token $j$ are given by:
\begin{equation}
    \alpha = \frac{\langle Q, K^T \rangle}{\sqrt{d_k}},
\end{equation}
and the attended HCT token representation is
\begin{equation}
    \tilde{F}_{\text{attn}} = \text{softmax}(\alpha) V.
\end{equation}

With $h$ heads, we compute head-specific outputs $\tilde{F}_{\text{attn}}^{(1)}, \dots, \tilde{F}_{\text{attn}}^{(h)}$, concatenate them along the feature dimension, and apply a final linear projection:
\begin{equation}
    F_{\text{fused}} = \text{Concat}\bigl(\tilde{F}_{\text{attn}}^{(1)}, \dots, \tilde{F}_{\text{attn}}^{(h)}\bigr) W_O \in \mathbb{R}^{1 \times d_f},
\end{equation}
where $W_O \in \mathbb{R}^{h d_v \times d_f}$.

\paragraph{Fusion for downstream processing.}
The fused token $F_{\text{fused}}$ captures how metadata reweights and contextualizes HCT features, effectively conditioning imaging-derived representations on patient-specific clinical information. In the subsequent MLP-Mixer module, we operate on $F_{\text{fused}}$ to model channelwise
interactions, and then apply global average pooling to obtain a single vector in $\mathbb{R}^{d_f}$ that is fed into the final classification head (described in Section~\ref{sec:feature_mixing_classification}).

\subsection{Feature Mixing and Classification}
\label{sec:feature_mixing_classification}
Given the fused token  $F_{\text{fused}} \in \mathbb{R}^{1 \times d_f}$ from the cross-attention module, we employ an MLP-Mixer to further refine the multimodal representation before classification. The MLP-Mixer operates on channel grids and consists of a stack of mixer blocks. Each block does channel-mixing, which mix information across channels for the token, enabling non-linear feature interactions within each token.

Formally, a single mixer block takes as input $U \in \mathbb{R}^{1 \times d_f}$ and computes:
\begin{align}
    U^{\text{ch}}  &= U + \text{MLP}_{\text{ch}}\bigl(\text{LayerNorm}(U)\bigr),
\end{align}
where $\text{MLP}_{\text{ch}}$ is applied along the channel dimension. Stacking multiple such blocks yields the refined feature sequence:
\begin{equation}
    F_{\text{mix}} = h_{\text{mix}}\bigl(F_{\text{fused}}\bigr) \in \mathbb{R}^{1 \times d_f}.
\end{equation}

To obtain a patient-level representation, we apply global average pooling over the token dimension:
\begin{equation}
    F_{\text{GAP}} = F_{\text{mix}}[1,:] \in \mathbb{R}^{d_f}.
\end{equation}

The pooled vector $F_{\text{GAP}}$ is then passed through a fully connected classification head:
\begin{equation}
    \hat{y} = \sigma\bigl(W_{\text{cls}} F_{\text{GAP}} + b_{\text{cls}}\bigr),
\end{equation}
where $W_{\text{cls}} \in \mathbb{R}^{1 \times d_f}$ and $b_{\text{cls}} \in \mathbb{R}$ are learnable parameters, $\hat{y} \in [0,1]$ denotes the predicted probability of edema, and $\sigma(\cdot)$ is the logistic sigmoid function. At inference time, a binary label is obtained by thresholding $\hat{y}$ (e.g., at 0.5).

\subsection{Training Objective}
We train AttentionMixer end-to-end using a binary cross-entropy loss over the dataset $\mathcal{D} = \{(X_i, M_i, y_i)\}_{i=1}^{N}$, where $y_i \in \{0,1\}$ is the ground-truth edema label and $\hat{y}_i$ is the corresponding predicted probability:
\begin{equation}
    \mathcal{L} = - \frac{1}{N} \sum_{i=1}^{N} \left[
y_i \log \hat{y}_i + (1 - y_i) \log (1 - \hat{y}_i)
\right].
\end{equation}

All model parameters (including metadata embedding network, cross-attention module, MLP-Mixer, and classification head) are optimized jointly using the Adamax optimizer with a learning rate of $10^{-3}$. Unless otherwise specified, we train for a fixed number of epochs with early stopping based on validation loss and monitor accuracy, F1-score, and AUC as primary evaluation metrics.

\section{Results}

We evaluate AttentionMixer on a curated cohort of patients (205 samples) with expert-annotated edema labels. All experiments use five-fold cross-validation at the patient level. For each fold, we compute accuracy, precision, F1-score, and area under the ROC curve (AUC), and report the mean values across folds.

\subsection{Experimental Setup and Implementation Details}

We partition the dataset into five cross-validation folds, ensuring that all scans from the same patient reside in a single fold to avoid information leakage. For each split, four folds are used for training and one for validation (for model selection and early stopping); we rotate the validation fold across the five runs and average metrics over all folds.

AttentionMixer is trained end-to-end using the binary cross-entropy loss described in Section~\ref{sec:feature_mixing_classification}, with the Adamax optimizer and a learning rate of $10^{-3}$. Unless otherwise specified, all components—including metadata embedding network, cross-attention module, MLP-Mixer, and classification head—are updated jointly. Hyperparameters such as batch size, number of epochs, and regularization strength are selected based on performance on the validation folds. We monitor accuracy, F1-score, and AUC during training and select the checkpoint with the best validation AUC for final evaluation in each cross-validation fold.

\subsection{Overall Performance}

Table~\ref{tab:main_comparison} summarizes the performance of AttentionMixer compared to several baselines, including unimodal models (HCT-only and metadata-only) and a prior multimodal fusion method (HyperFusion). Among HCT-only models, the 3D CNN and MLP-Mixer backbones achieve accuracies of 78.68\% and 74.63\%, respectively, while a vanilla ViT underperforms on this dataset (accuracy 70.24\%, F1-score 55.92\%). These results suggest that, in the absence of explicit multimodal conditioning, standard vision backbones struggle to fully capture edema-relevant patterns from HCT alone.

The metadata-only MLP \cite{haykin1994neural} baseline attains an accuracy of 84.88\%, precision of 89.90\%, F1-score of 82.77\%, and AUC of 92.31\%, indicating that routine clinical variables already carry substantial predictive information for edema status. HyperFusion, which modulates CNN parameters based on metadata, does not outperform the metadata-only baseline in this setting (accuracy 66.66\%, AUC 86.66\%), suggesting that naïve or overly complex parameter modulation can be fragile when sample sizes are modest.

In contrast, AttentionMixer—combining a ViT-based HCT encoder, cross-attention fusion, and MLP-Mixer refinement—achieves an average accuracy of 87.32\%, precision of 92.10\%, F1-score of 85.37\%, and AUC of 94.14\%. These results demonstrate the effectiveness of integrating structural imaging and clinical metadata for edema classification and show that (i) integrating HCT with clinical metadata yields clear gains over HCT-only models, (ii) a structured, attention-based fusion strategy can outperform both unimodal baselines and a strong multimodal comparator, and (iii) the proposed architecture leverages both modalities to improve patient-level edema detection beyond what either modality achieves alone.

\subsection{Ablation Study on Fusion and Refinement}

To assess the impact of our fusion and refinement mechanisms, we conduct an ablation study comparing different variants of AttentionMixer and a simple early-fusion baseline. The results are summarized in Table~\ref{tab:ablation_fusion}.

We consider the following configurations:
\begin{itemize}
    \item AttentionMixer (w/o CA): the full model with cross-attention removed, so HCT and metadata features are combined without attention-based weighting, while the MLP-Mixer refinement is retained.
    \item AttentionMixer (w/o MLP-Mixer): the full model with cross-attention fusion but without the MLP-Mixer; fused features are pooled and passed directly to the classifier.
    \item Early Fusion: a baseline where HCT and metadata features are concatenated and processed without explicit attention or mixer-based refinement.
    \item Only MLP-Mixer: a degenerate variant that uses an MLP-Mixer alone without the proposed ViT-based encoder and cross-modal fusion (included as a lower bound).
\end{itemize}

Relative to these variants, the full model (Table~\ref{tab:main_comparison}) achieves the highest AUC (94.14\%). Removing cross-attention (AttentionMixer w/o CA) yields a lower AUC (92.66\%) and  accuracy. Removing the MLP-Mixer (AttentionMixer w/o MLP-Mixer) also leads to a drop in accuracy and F1-score compared to the full model, even though AUC remains strong (93.35\%), suggesting that lightweight post-fusion refinement is beneficial even after attention-based fusion.

The Early Fusion baseline performs competitively but consistently worse than the full AttentionMixer across all metrics, with a notably lower AUC of 90.00\%. The Only MLP-Mixer configuration performs poorly across metrics, confirming that both the self-supervised ViT encoder and the cross-modal fusion mechanism are crucial for effective edema classification. Taken together, these ablations support our design choices: cross-attention is important for principled multimodal fusion, and the MLP-Mixer provides additional gains through parameter-efficient refinement of the fused representation.

\subsection{ROC Curve Analysis}

\begin{figure}[!ht]
    \centering
    \includegraphics[width=0.45\textwidth]{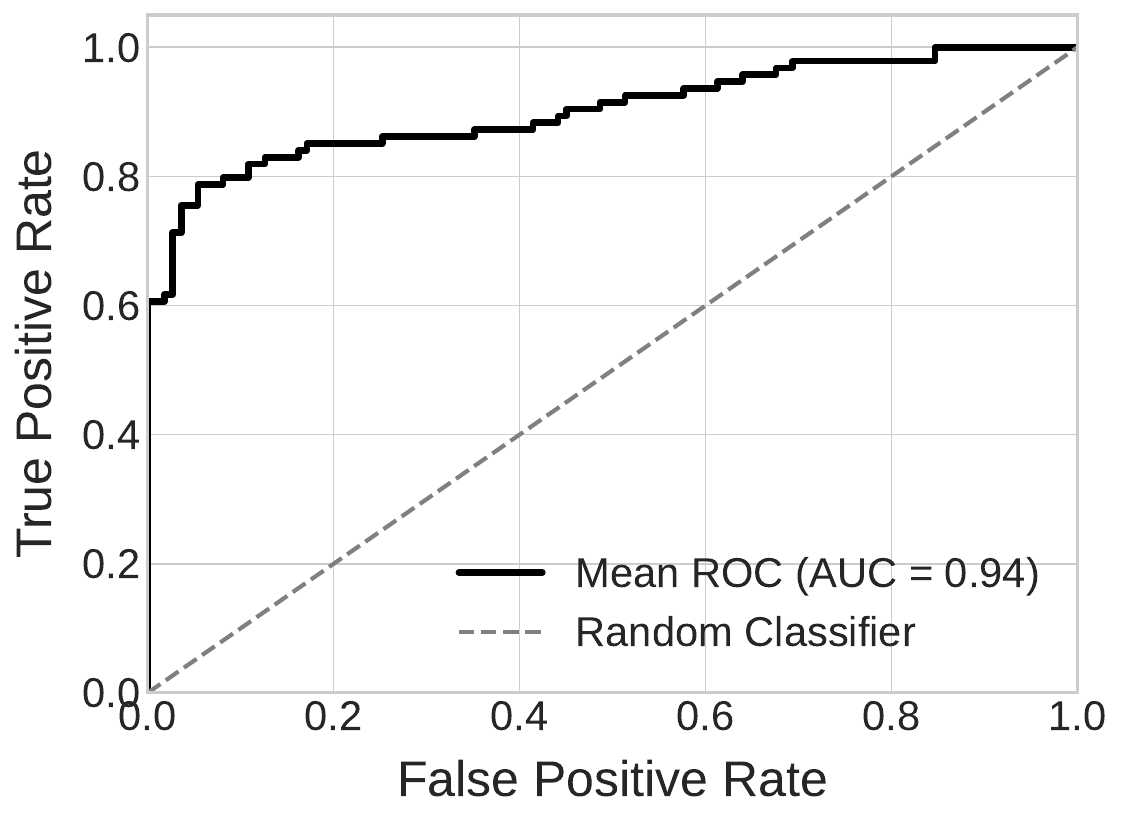}
    \caption{Average ROC curve of AttentionMixer across 5-fold cross-validation. The mean AUC is 94\%.}
    \label{fig:roc_curve_avg}
\end{figure}

Figure~\ref{fig:roc_curve_avg} shows the average ROC curve of AttentionMixer across the five cross-validation folds. The model achieves a mean AUC of 94\%, consistent with Table~\ref{tab:main_comparison}, indicating excellent discriminative capability in distinguishing between edematous and non-edematous cases. The ROC curve lies well above the diagonal across the entire range of false-positive rates, suggesting that AttentionMixer can be tuned to operate at clinically meaningful operating points—for example, prioritizing high sensitivity for screening or higher specificity when minimizing false positives is critical.

\subsection{Metadata Feature Importance}
To assess the contribution of individual clinical variables to multimodal edema classification, we perform a permutation feature importance analysis across the validation sets of all folds. Understanding metadata importance is critical in multimodal learning, as it reveals which clinical factors the model depends on when refining HCT-derived features through cross-attention. Unlike traditional feature importance measures that rely solely on model weights, permutation-based evaluation directly quantifies the impact of each feature on predictive performance.

For each metadata variable, its values are randomly permuted across patients while all other modalities and feature dimensions are kept intact. This operation disrupts the association between that feature and the outcome without altering the marginal distribution of the data. The model is then evaluated on the permuted validation set, and the corresponding drop in AUC ($\Delta \text{AUC}$) is recorded. A larger $\Delta \text{AUC}$ indicates that the model relies more heavily on that variable for accurate prediction. This process is repeated independently for every metadata feature and for every fold of cross-validation. The importance scores are then averaged across the five folds to obtain a stable and statistically robust estimate of feature relevance.

This analysis provides direct interpretability for the multimodal fusion mechanism. Since metadata embeddings serve as keys and values in the cross-attention module, important metadata features exert greater influence on how HCT features are re-weighted and refined. By observing which permutations most degrade model performance, we gain insight into the clinical variables that meaningfully shape the fused representation and, ultimately, the edema predictions.

\begin{figure}[!t]
    \centering
    \includegraphics[width=\linewidth]{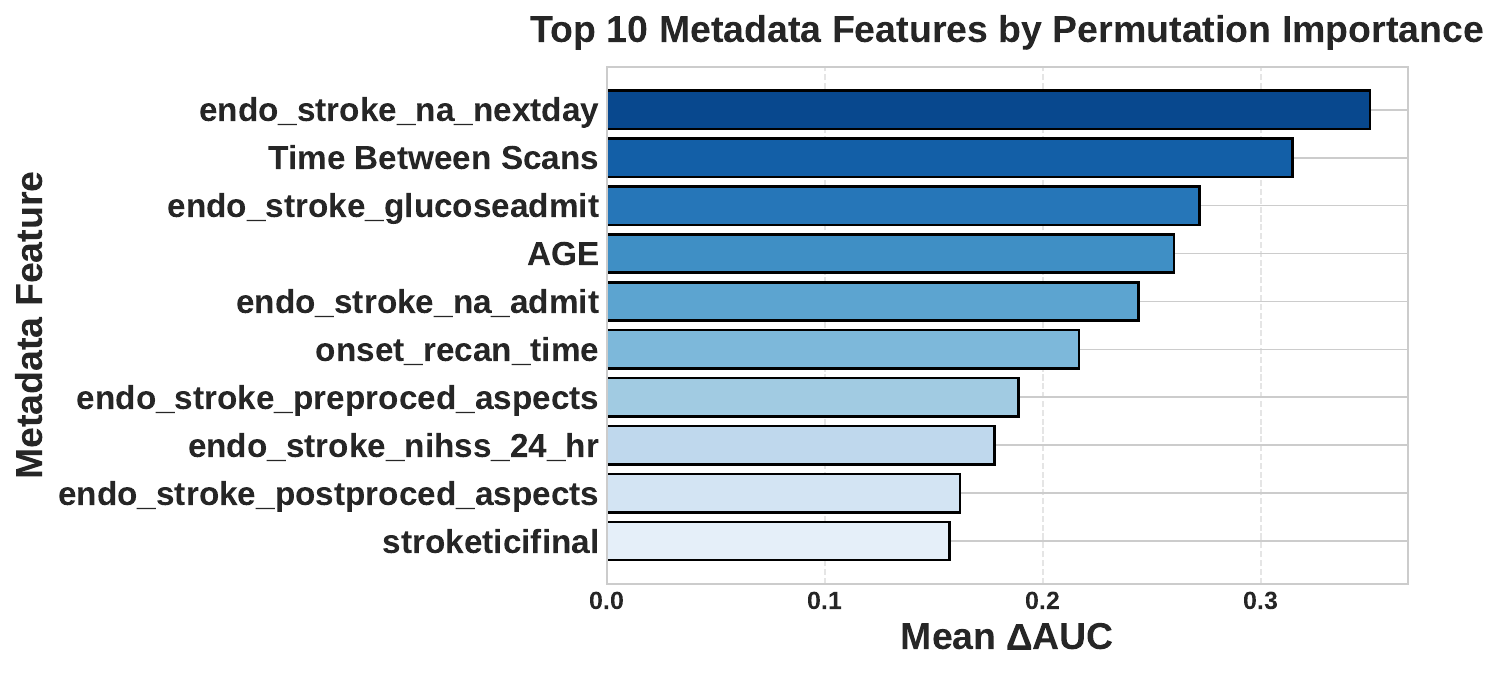}
    \caption{
        Top-10 metadata variables ranked by permutation feature importance. 
        Importance is measured as the mean decrease in AUC ($\Delta \text{AUC}$) across five 
        cross-validation folds when each feature is independently permuted. 
        Higher values indicate stronger contribution of the metadata feature to 
        the multimodal edema classification model.
    }
    \label{fig:metadata_importance}
\end{figure}

Figure~\ref{fig:metadata_importance} illustrates the top 10 metadata variables ranked by their mean $\Delta \text{AUC}$ across folds. Variables with larger decreases in AUC when permuted contribute more strongly to the multimodal model, highlighting clinically plausible factors that AttentionMixer uses in conjunction with HCT to detect edema. Qualitatively, we observe that variables such as sodium next day, glucose on admission, time between scans, and age tend to produce larger drops in AUC when permuted, which aligns with clinical intuition about factors associated with edema burden and severity.

\subsection{Distribution of Predicted Probabilities}
\begin{figure}[!t]
    \centering
    \includegraphics[width=\linewidth]{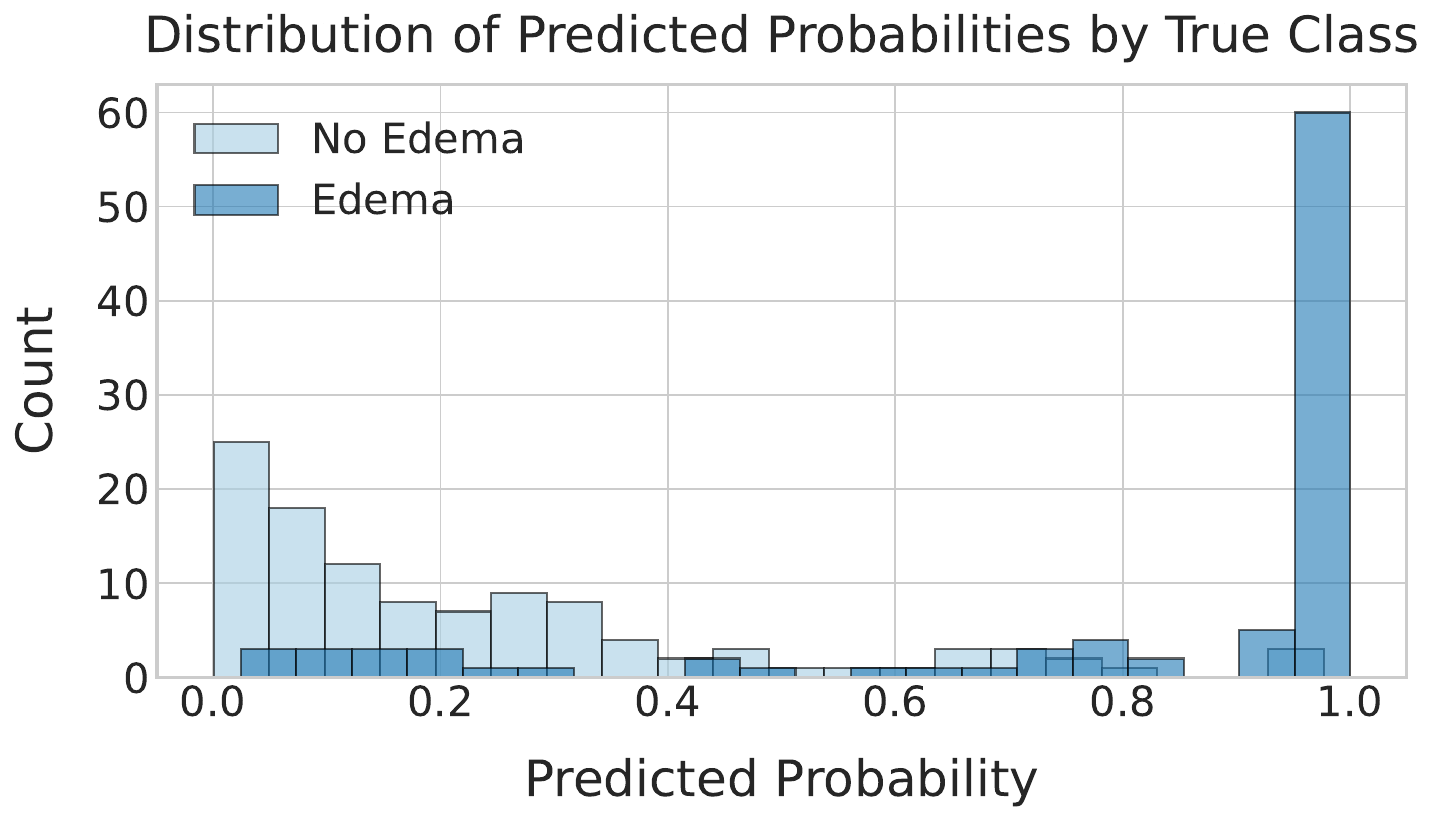}
    \caption{
        Distribution of predicted probabilities for edema and non-edema cases. 
    }
    \label{fig:probability_distribution}
\end{figure}

To better understand model confidence and error behavior, we examined the distribution of predicted edema probabilities stratified by the ground-truth label. Figure~\ref{fig:probability_distribution} shows overlaid histograms of predicted probabilities for non-edema and edema cases across all validation folds.

The distributions of two classes exhibit markedly different shapes. Predictions for edema cases are highly concentrated near 1.0, forming a sharp peak that reflects strong and consistent 
model confidence. In contrast, non-edema predictions span a much broader range, with several local modes corresponding to varying levels of uncertainty.

\section{Limitations and Future Work}

While the proposed \emph{AttentionMixer} framework shows promising performance for multimodal edema detection, several limitations warrant discussion and point toward future research directions.

\textit{Single-center, retrospective cohort.}
Our experiments are conducted on a single curated dataset with retrospective labels, which may limit generalizability across institutions, scanners, and acquisition protocols. In particular, differences in HCT sequences, vendor-specific reconstruction, and patient populations could affect model performance. Future work will include multi-center, multi-scanner external validation and domain adaptation strategies to assess and improve robustness to distribution shifts.

\textit{Binary classification rather than fine-grained edema assessment.}
AttentionMixer is trained for binary edema detection (presence vs.\ absence) at the patient level, without explicitly modeling edema volume, spatial extent, or subtype (e.g., vasogenic vs.\ cytotoxic). Although this formulation is clinically meaningful as a screening or triage tool, it does not directly provide quantitative measurements needed for detailed treatment planning. An important extension is to couple the current classifier with a segmentation head or regression module that predicts edema burden and progression over time.

\textit{Limited modality and sequence diversity.}
In this work, we focus on structural HCT and a fixed set of tabular clinical variables. Many clinical workflows also incorporate additional modalities, such as diffusion-weighted imaging, perfusion, susceptibility-weighted imaging, or CT in mixed cohorts. Moreover, not all sequences are consistently available for every patient. Extending AttentionMixer to handle heterogeneous multi-sequence inputs and to operate robustly under partial modality availability (e.g., missing sequences at test time) is a natural next step.

\textit{Label quality and cohort size.}
Although edema labels are curated by experts, they remain subject to inter-rater variability and potential labeling noise, especially in borderline cases. The cohort size is also modest relative to the capacity of modern deep models. While we mitigate these issues via self-supervised pretraining, future work should incorporate larger, more diverse cohorts and explore techniques that explicitly model label uncertainty, such as probabilistic labeling, label smoothing, or noise-robust loss functions.

\textit{Calibration, fairness, and deployment considerations.}
We primarily evaluate discrimination metrics (accuracy, F1-score, AUC) and do not comprehensively assess probability calibration or fairness across demographic subgroups. For clinical deployment, it is important to ensure that predicted probabilities are well calibrated and that performance is consistent across age, sex, and other relevant strata. Future work will therefore include calibration analysis, subgroup performance evaluation, and user-centered studies on how to integrate AttentionMixer into radiology workflows (e.g., as a PACS-integrated decision-support tool) while preserving transparency and clinician control.

\textit{Interpretability and human–AI interaction.}
While cross-attention weights and permutation-based feature importance offer some interpretability, they do not fully capture the complexity of multimodal decision-making. Further work is needed to develop richer explanations that jointly highlight salient image regions and influential metadata, for example via attention roll-outs, counterfactual analysis, or concept-based explanations. These tools could facilitate better human–AI interaction, enabling clinicians to vet and trust model recommendations.

\section{Conclusion}
We introduced \emph{AttentionMixer}, a unified multimodal deep learning framework for patient-level detection of brain edema that jointly leverages 3D HCT and structured clinical metadata. The architecture combines a self-supervised ViT-based HCT encoder, a cross-attention module that conditions imaging features on patient-specific metadata, and an MLP-Mixer for parameter-efficient post-fusion refinement. To address real-world clinical constraints, AttentionMixer includes a learnable embedding for entirely missing metadata, enabling robust training and inference without discarding patients with incomplete records.

On a curated cohort with expert edema annotations, AttentionMixer outperforms strong HCT-only, metadata-only, and multimodal baselines, achieving higher accuracy, F1-score, and AUC. Ablation studies demonstrate that both cross-attention and the MLP-Mixer contribute meaningfully to performance, and permutation-based feature importance analysis reveals clinically plausible metadata variables that drive the model’s decisions. Together, these findings support the central premise of this work: principled, interpretable fusion of structural imaging and clinical context can substantially enhance automated edema detection.

Looking ahead, we envision AttentionMixer as a building block for more comprehensive neuroimaging decision-support systems that integrate additional imaging modalities, model edema burden and progression, and are validated across institutions and patient populations. By aligning multimodal deep learning more closely with how clinicians naturally synthesize imaging and clinical information, such systems have the potential to improve the reliability, efficiency, and transparency of AI-assisted care in acute and chronic neurologic disease.
\printcredits

\bibliographystyle{cas-model2-names}
\bibliography{bib} 

\end{document}